\documentclass{llncs}
\usepackage{makeidx}  
\usepackage{lipsum}
\usepackage{multirow}
\usepackage[usenames, dvipsnames]{color}
\usepackage{graphicx}
\usepackage{wrapfig}
\usepackage{url}
\usepackage{hhline}
\usepackage{caption}
\usepackage{cleveref}
\usepackage{soul}
\usepackage{hyperref}

\begin{document}
\frontmatter          
\pagestyle{headings}  
%
%
\title{Sentiment Recognition in Egocentric Photostreams}
\titlerunning{Towards Sentiment Recognition Model}
\author{Estefania Talavera \inst{1,2} \and Nicola Strisciuglio\inst{1} 
\and Nicolai Petkov \inst{1} \and Petia Radeva \inst{2,3} }
\authorrunning{Estefania Talavera et al.}  
\tocauthor{Estefania Talavera Martinez, Nicola Strisciuglio,
Petia Radeva and Nicolai Petkov}
\institute{Intelligent Systems Group, University of Groningen, The Netherlands\\
\email{e.talavera.martinez@rug.nl},
\and
Department of Mathematics and Computer Science, University of Barcelona, Spain\\
\and
Computer Vision Center, Barcelona, Spain}

\maketitle              

\begin{abstract}
Lifelogging is a process of collecting rich source of information about daily life of people. In this paper, we introduce the problem of sentiment analysis in egocentric events focusing on the moments that compose the images recalling positive, neutral or negative feelings to the observer. We propose a method for the classification of the sentiments in egocentric pictures based on global and semantic image features extracted by Convolutional Neural Networks. We carried out experiments on an egocentric dataset, which we organized in 3 classes on the basis of the sentiment that is recalled to the user (positive, negative or neutral).
\end{abstract}

\section{Introduction}
\vspace{-0.5em}

Mental imagery is the process in which the feeling of an experience is imagined by a person in the absence of external stimuli. It has been assumed by therapists to be directly related with emotions \cite{Holmes2006PositiveMood}, opening some questions when images describing past moments of our lives are available: \textit{Can an image make the process of mental imagery easier?} or \textit{Can specific images help us to retrieve or imply feelings and moods?} 

Lifelogging is a recent trend consisting in constructing a digital collection from an egocentric point of view of the events of a person that wears a recording device. It is a tool for the analysis of the lifestyle of users, since it provides objective information of what happened during different moments of the day, and a powerful tool for memory enhancement \cite{Lee2008LifeloggingImpairment}. Using wearable cameras, each day up to 2000 egocentric photostreams are usually recorded, i.e. up to 70000 per month. A lot of these images are redundant, non-informative or routine and thus without special value for the wearer to be preserved. Usually, users are interested in keeping special moments, images with sentiments that will allow them in the future to re-live the personal moments captured by the camera. An automatic tool for sentiment analysis of egocentric images is of high interest to make possible the processing of the big collection of lifelogging data and keeping out just the images of interest i.e. of high charge of positive sentiments.
\begin{figure}[ht!]
\centering
\includegraphics[width=1\textwidth,height=25mm]{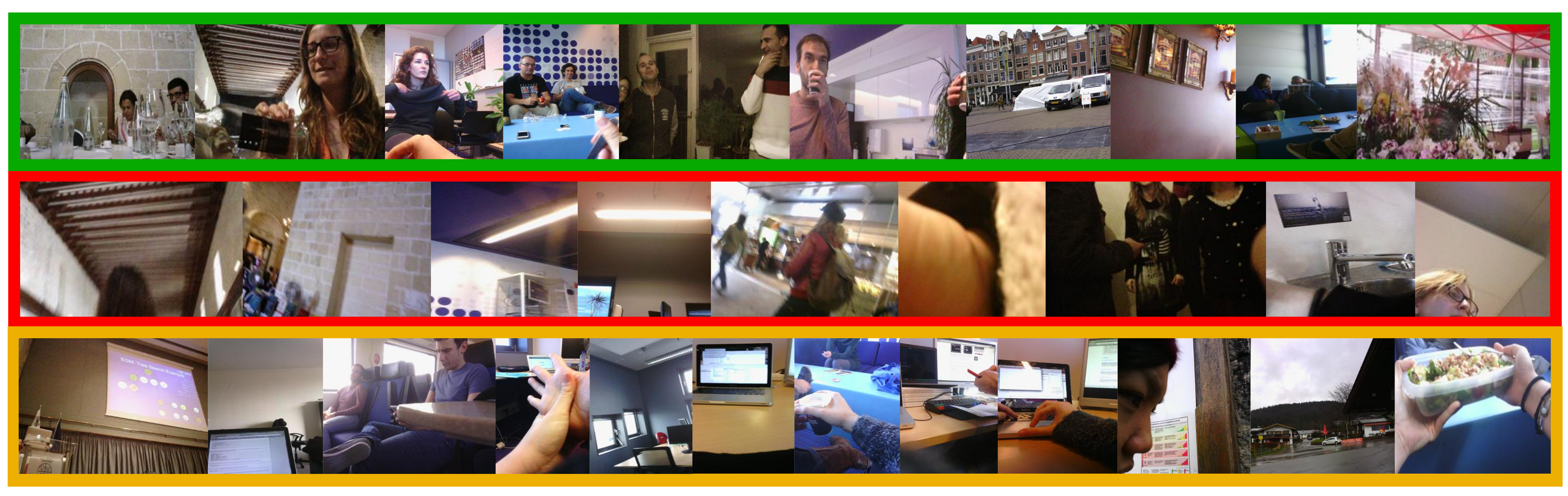}
\vspace{-1.5em}
\caption{ Examples of Positive (green), Negative (red) and Neutral (yellow) images.}
\label{fig:exampleImages}
\vspace{-2.5em}
\end{figure}

However, the automatic sentiment image analysis is a complicated task first of all, because of the lack of clear definition of it. There is no consensus between  the different sentiment ontologies  in the literature. Table \ref{table:datasets} illustrates the ambiguity of the problem, reporting several sentiments ontology in images. The first group \cite{Machajdik2010AffecitveTheory,You2016BuildingBenchmark,Yi2014LearningScratch} assigns 8 main sentiments as excitement, awe or sadness to the images with assigned discrete positive (1) and negative (-1) sentiment value. The second group \cite{Dan-Glauser2011TheSignificance,Lang1997InternationalRatings} defines a different set of sentiments as valence or arousal and discrete positive (1), neutral (0) or negative (-1) values assigned to the images according to the sentiments. In contrast, the third group \cite{Nojavanasghar2016EmoReact:Children} assigns up to 17 sentiments (6 basics and 9 complex) and each image of the dataset is assigned a continuous value in a scale from 1 to 4. Given the ambiguity of the semantic sentiment assignment, with labels difficult to classify into positive or negative sentiments, the last group \cite{Borth2013Large-scalePairs} defines up to 3244 Adjective Noun Pairs (ANP) (e.g. 'beautiful\_girl') and assigns a continuous sentiment value in a range of [-2,2] to them. The main idea is that the same object according to its appearance has positive or negative sentiment value like \textit{'angry\_dog' (-1.55)} and \textit{'adorable\_dog' (+1.45)}. A natural question is until which extent the 3244 ANPs represent a scene captured by the image, taking into account the difficulty to detect them automatically (Mean average accuracy $\sim$25\%). 

\begin{table}[ht!]
\vspace{-3em}
\centering
\caption{Different image sentiment ontologies.}
\label{table:datasets}
\begin{tabular}{c|c|c|c|c}
DataSets & Source & Images & Semantic sentiment labels & \begin{tabular}[c]{@{}c@{}}Sentiment\\ Values \end{tabular} \\ \hhline{|=|=|=|=|=|}
\begin{tabular}[c]{@{}c@{}}Abstract 
\\ \&Artphoto \\ \cite{Machajdik2010AffecitveTheory} 
\end{tabular} &  & \begin{tabular}[c]{@{}c@{}}280\\\& 806\end{tabular}& \begin{tabular}[c]{@{}c@{}}positive: contentment, amusement, \\excitement, awe, \\ negative: sadness, fear, \\ disgust, and anger\end{tabular} & \{1,-1\} \\ \hline
\begin{tabular}[c]{@{}c@{}}You's Dataset\\ \cite{You2016BuildingBenchmark}\end{tabular} &  \begin{tabular}[c]{@{}c@{}}Flickr\\ Instagr\end{tabular} & 23000 & \begin{tabular}[c]{@{}c@{}}positive: contentment, amusement, \\ excitement,  awe,\\ negative: sadness, fear,\\ disgust, and anger\end{tabular} &  \{1,-1\} \\ \hline
\begin{tabular}[c]{@{}c@{}}CASIA-WebFace\\ \cite{Yi2014LearningScratch}\end{tabular}
 &  & 494k & \begin{tabular}[c]{@{}c@{}}anger, disgust, fear \\ happy, neutral, sad, surprise\end{tabular} & [1,0,-1] \\  \hline
IAPS\cite{Lang1997InternationalRatings} &  & 1182 & valence, arousal, and dominance & [1,7] \\  \hline
GAPED \cite{Dan-Glauser2011TheSignificance} &  & 732 & \begin{tabular}[c]{@{}c@{}}valence, arousal, \\ and normative significance\end{tabular} & \{1,0,-1\} 
\\ \hline
\begin{tabular}[c]{@{}l@{}}EmoReact\\ \cite{Nojavanasghar2016EmoReact:Children}\end{tabular} 
& Youtube & \begin{tabular}[c]{@{}c@{}}1102\\clips\end{tabular} & \begin{tabular}[c]{@{}c@{}}17 sentiments: 6 basic emotions \\ (positive: happiness, surprise, \\ negative: sadness, fear,\\ disgust, and anger),\\ and 9 complex emotions:  (curiosity, \\ uncertainty, excitement,\\ attentiveness, exploration, confusion, \\ anxiety, embarrassment, frustration).\end{tabular} & [1,4] 
\\ \hline
\begin{tabular}[c]{@{}l@{}}VSO +\\ TwitterIm\cite{Borth2013Large-scalePairs}\end{tabular} & \begin{tabular}[c]{@{}c@{}}Flickr\\ Twitter\end{tabular} & 0.5M & \begin{tabular}[c]{@{}c@{}}Not, but Adjective Noun\\ Pairs (3244) \end{tabular}  & \begin{tabular}[c]{@{}c@{}}\small{Flickr}[-2,2]\\ \small{Twitter}[-1,1]  \end{tabular} \\ \hline 
\begin{tabular}[c]{@{}l@{}}
You\_RobustSet\\ \cite{You2015RobustNetworks}\end{tabular} & \multicolumn{1}{c|}{Twitter} &  \multicolumn{1}{c|}{1269}  & \begin{tabular}[c]{@{}l@{}}
Non-semantic labels:\\ Positive and  Negative \end{tabular} & \{1,-1\} \\ \hline 
\begin{tabular}[c]{@{}l@{}}
UBRUG-\\EgoSenti*\end{tabular} & \multicolumn{1}{l|}{\begin{tabular}[c]{@{}c@{}}Wearable\\ Camera\end{tabular}} &  \multicolumn{1}{c|}{12088}  & \begin{tabular}[c]{@{}l@{}}
Non-semantic labels:\\ Positive, Neutral and Negative \end{tabular} & \{1,0,-1\}
\end{tabular}
\vspace{-2.6em}
\end{table}

Given the difficulty of image sentiment determination, ambiguity and lack of consensus in the bibliography, added by the difficulty of the egocentric images, we focus on the image sentiment as a discrete value expressing a ternary sentiment value (positive (1), negative (-1) or neutral (0) value) similar to \cite{You2015RobustNetworks}. Egocentric data is of special difficulty, since we do not observe the wearer and his/her, i.e. from facial or corporal expressions, but rather from the perspective of what the user sees. Moreover, in real life fortunately, negative emotions have much less prevalence than neutral and positive, that makes very difficult to have enough examples of negative egocentric images and events. Thus, the problem we address in this article is what effect an egocentric image or event has on an observer (positive, neutral or negative) (see Fig.\ref{fig:exampleImages}), instead of attempting to specify an explicit semantic image sentiment like sadness; and how to develop automatic tool for sentiment value detection (positive, vs. neutral vs. negative)  and egocentric dataset in order to validate its results. Going further, in contrast to the published work, we claim to automatically analyse the sentiment value of egocentric events i.e. a group of sequential images that represents the same scene. In the case of egocentric images, the probability that a single image describes an event is low; there are a lot of images that just capture wall, sky, ground or partially objects.  For this reason, we are interested to automatically discover how the event captured by the camera influences the observer, that is to automatically determine the  ternary sentiment values of the events, which are richer in information and involve the whole moment's experience. For example, an event being in a dark and narrow, grey space would influence negatively, a routine scene like working in the wearer's office could influence the observer neutrally and 
an event where the wearer has spent some time with friends in a nice outdoor space could influence positively to the observer.

Automatic sentiment analysis from images is a recent research field. In the literature, sentiment recognition in conventional images has been approached by computing and combining visual, textual, or audio features \cite{Nojavanasghar2016EmoReact:Children,Poria2014FusingContent,Wang2014MicroblogModel,You2016Cross-modalityMultimedia}. Other characteristics, such as facial expressions have also been used for sentiment prediction \cite{Yuan2013SentributeDescriptors}. The combination of visual and textual features extracted from images is possible due to the wide use of online social media and microblogs, where images are posted accompanied by short comments. Therefore, multimodal approaches were proposed, where both sources of information are merged \cite{Wang2014MicroblogModel,You2016Cross-modalityMultimedia} for automatic sentiment value detection. 

Recently, with the outstanding performance of Convolutional Neural Networks (CNN), several approaches to sentiment analysis relied on deep learning techniques for classification and/or features extraction combined with other networks or methods \cite{Campos2015DivingPrediction,Levi2015EmotionPatterns,You2016BuildingBenchmark,Yu2016VisualNetworks}. The work in \cite{You2016BuildingBenchmark} applies fine-tuning on the AlexNet to classify the 8 emotions: sadness, angry, content, etc. 
In contrast, in \cite{Campos2015DivingPrediction} they propose to fine-tuned CaffeNet with oversampling to classify into Positive or Negative sentiments. In \cite{Levi2015EmotionPatterns} a novel transformations of image intensities to 3D spaces is proposed to reduce the amount of data required to effectively train deep CNN models. In \cite{Yu2016VisualNetworks} the authors use logistic regression to classify into 3 sentiments using CNN features. In \cite{Chen2014DeepSentiBank:Networks}, the authors perform a fine-tuning on a CNN model and modify the last layer to classify 2089 ANPs. However, no work has addressed the sentiment image and event analysis in egocentric datasets.

To address the egocentric data sentiment analysis, we propose to combine semantic concepts in terms of ANPs, given that they have sentiment values associated \cite{Borth2013Large-scalePairs}, with general visual features extracted from a CNN \cite{Krizhevsky2012ImageNetNetworks}. ANPs represent a finite subset of concepts present in the image, so they bring strong sentiment value, but can not ensure to cover the whole image content. Visual features extracted by CNNs can help to summarize the whole image content in an intermediate level. We  test our method on a new egocentric dataset of 12088 pictures with ternary sentiment values acquired from 3 users and 20 days. A very preliminary stage of this work has been presented in \cite{Talavera2017TowardsAnalysis}. 

Therefore, our contributions here are three-fold: a) a model for ternary sentiment value analysis in egocentric images, b) extension of the approach to egocentric events, and c) the first egocentric sentiment value dataset from 12088 images covering 20 days of 3 persons.

The paper is organized as follows. We describe the proposed approach and the dataset in Sections 2 and 3, respectively. In Section \ref{sec:experimentsetup}, we describe the experimental setup, the quantitative and qualitative evaluation, and discuss our findings. Finally, Section \ref{sec:conclusions} draws conclusions and outlines future works.

\vspace{-1em}
\section{Proposed Method}
\label{sec:modelapproach}
\begin{figure}[ht!]
\vspace{-1.5em}
\centering
\includegraphics[width=0.8\textwidth,height=40mm]{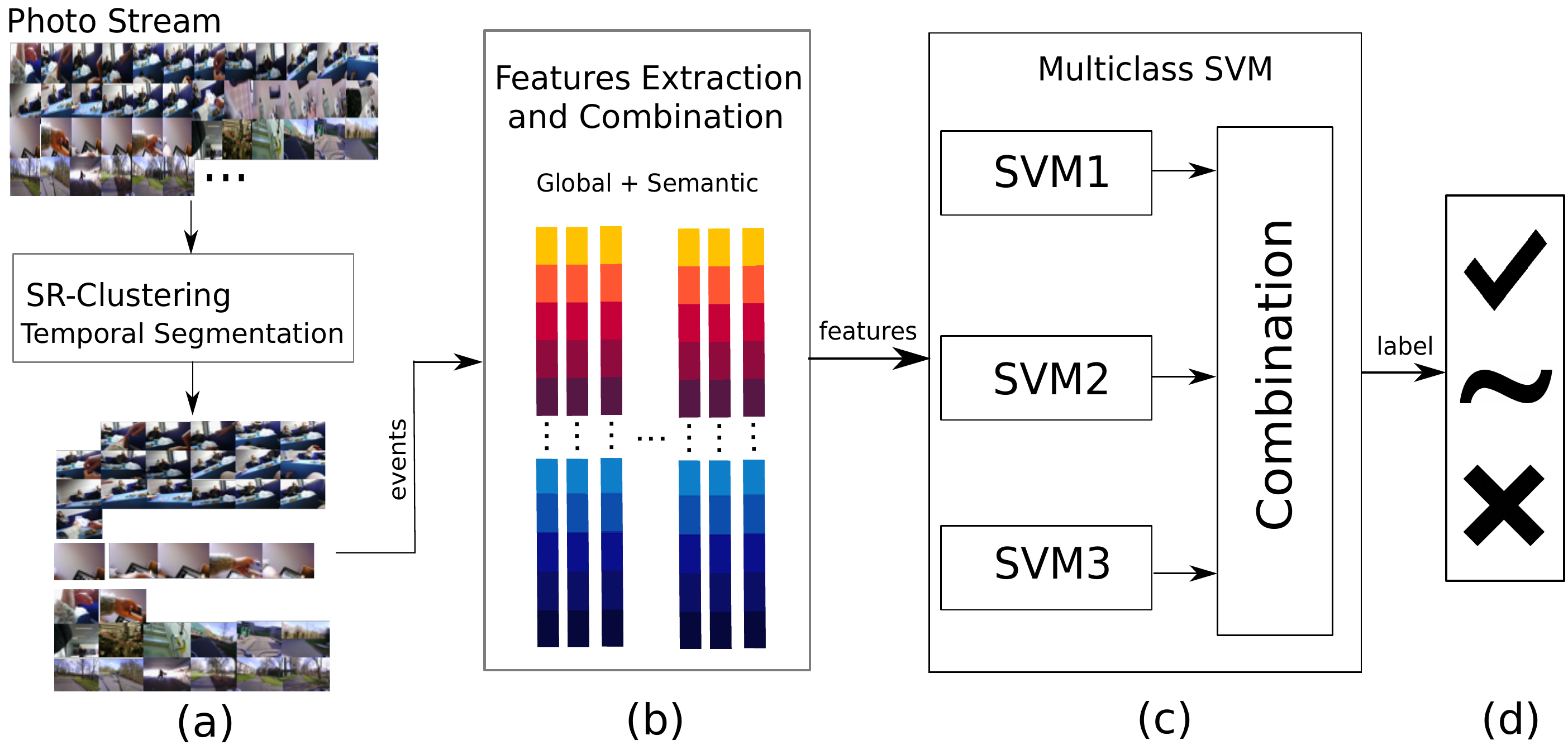} 
\vspace{-0.5em}
\caption{Architecture of the proposed method. (a) Temporal segmentation of the photostream into events. (b) CNN and ANPs features are extracted from the images and (c) used as input to the trained multi-class SVM model. (d) The model labels the input image as Positive, Neutral or Negative.}
\label{fig:methodproposed}
\vspace{-3em}
\end{figure}

\vspace{-0.5em}
In this section, we describe the proposed method for sentiment recognition from egocentric photostreams, which is based on visual (extracted by CNN) and semantic (in terms of ANPs) features extracted from the images. An architectural overview of the proposed system is depicted in Fig. \ref{fig:methodproposed}.

\vspace{-0.5em}
\subsubsection{a) Temporal Segmentation:}
\label{temporalsegmentation}
\vspace{-0.5em}

Given that egocentric images have smaller field of view and thus do not capture entirely  the context of the event, we need to detect the events of the days. To this aim, we apply the SR-Clustering algorithm for temporal segmentation of photostreams \cite{Dimiccoli2015SR-Clustering:Segmentation}. The clustering procedure is performed on an image representation that combines visual features extracted by a CNN with semantic features in terms of visual concepts extracted by Imagga's auto-tagging technology (http://www.imagga.com/solutions/auto-tagging.html).

\vspace{-0.5em}
\subsubsection{b) Features Extraction:}
\vspace{-0.5em}

For the computation of the semantic features in terms of the ANPs, we use the DeepSentiBank Network \cite{Chen2014DeepSentiBank:Networks}. Given an image, the DeepSentiBank network considers the 2089 best performing ANPs. Applying the DeepSentiBank on them gives  a 2089-D feature vector, where the feature values correspond to the ANPs likelihood in the image. These values are multiplied by the sentiment value associated to the concepts. Note that each ANP has a positive or negative sentiment value assigned, but not 0 for a neutral sentiment.

However, the 2089 ANPs not necessarily have the power to explain the "richness" of any scene in an image. Hence, we integrate the ANPs feature vector with a feature descriptor provided by the penultimate layer if a CNN \cite{Krizhevsky2012ImageNetNetworks} that summarizes the whole context of the image. The resulting feature vector is composed by 4096 features. 
We combine the ANPs and the CNN feature vectors into a 6185-D feature vector, in order to construct a more reliable and rich image representation that relates image semantics expressed by the ANPs with clear sentiment value with the CNN cues as an intermediate image representation. We apply the Signed Root Normalization (SRN) to transform the CNN feature vectors to a more uniformly distributed space followed by a $l_2$-normalization \cite{Zheng2014SeeingEvidences}. 

\vspace{-0.5em}
\subsubsection{c) Classification:} 
\vspace{-0.5em}
We use the proposed feature vectors to train a multi-class SVM classifier due to its high generalization capability \cite{Joachims2000EstimatingEfficiently}. This is ensured by the SVM learning algorithm that finds a separation hyperplane that maximizes the separation margin between the classes. 
We employ a 1-vs-all design for the multi-class problem, as suggested in \cite{Foggia2015ReliableEnvironments}. 
The cardinality of the classes in the proposed dataset is not balanced, which affects the computation of the training error cos
In order to classify an event, we use a majority vote on the image level classification output. 

\vspace{-1em}
\section{Dataset}
\label{sec:dataset}
\vspace{-0.5em}

We collected a dataset of 12471 egocentric pictures, which we call UBRUG-EgoSenti. The users were asked to wear a Narrative Clip Camera, which takes a picture every 30 seconds, hence each day around 1500 images are collected for processing. The images have a resolution of 5MP and JPG format.

We organize the images into events according to the output of the SR-clustering algorithm \cite{Dimiccoli2015SR-Clustering:Segmentation}. From the originally recorded data, we discarded those events that are composed of less than 6 images, so obtaining a dataset composed of 12088 images grouped in a total of 233 events, with an average of 51.87 images  per event and std of 52.19. We manually labelled the events following how the user felt while reviewing them by assigning \emph{Positive}, \emph{Negative} and \emph{Neutral} values to them, some examples of which are given in Fig \ref{fig:exampleImages}. The dataset, for which the details are in Table \ref{tabledataset}, is publicly available and can be downloaded from: \url{http://www.ub.edu/cvub/dataset/.} 
 \begin{table}[ht!]
 \centering
 \vspace{-2.5em}
 \caption{Description of the UBRUG-EgoSenti dataset.}
 \label{tabledataset}
 \setlength{\tabcolsep}{12pt}
 \begin{tabular}{c|c|c|c|c}
 Class    & Images & \#Events & Mean Im Event & Std Im Event \\ \hline
 Positive & 4737 & 83 & 57.07 & 52.34 \\
 Neutral  & 6169 & 107 & 57.65 & 57.18 \\
 Negative & 1182 & 43 & 27.49 & 26.44 \\ \hline
 Total    & 12088 & 233 & 51.88 & 52.19
 \end{tabular}
 \vspace{-2em}
 \end{table}
\vspace{-1.5em}
\section{Experiments}
\label{sec:experimentsetup}
\vspace{-0.8mm}
\subsection{Evaluation and results}
\vspace{-0.5mm}
We carried out 10-fold cross-validation.  Events from different classes are uniformly distributed among the various folds, which are thus independent from each other.
We evaluated the performance of the proposed system on single images and at event level. For the UBRUG-Senti dataset, the groundtruth labels are given at event level. All the images that compose a certain event, are considered as having the same label of such event. Given an event composed of $M$ images, we aggregate the $M$ classification decisions by majority vote.
We measure the performance results of our method by computing the average accuracy. 

\begin{figure}[ht!]
\vspace{-1.2em}
\centering
\includegraphics[width=1\textwidth,height=32mm]{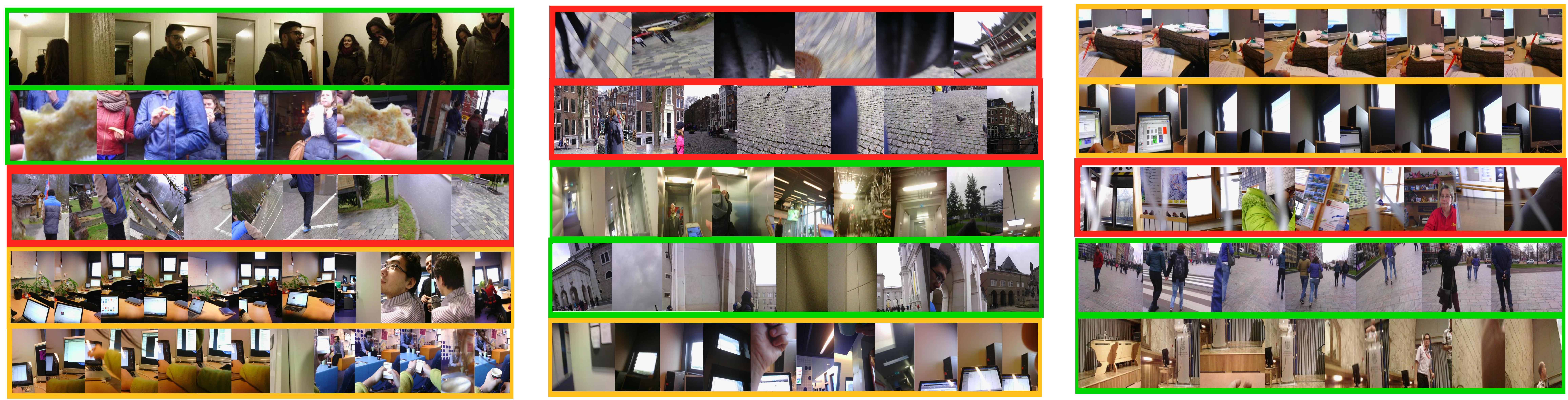}
\vspace{-2.2em}
\caption{Examples of the automatic event sentiment classification. The events are grouped based on the sentiment defined by the user: (right) Positive, (middle) Negative, and (left) Neutral. The events frame colour corresponds to the label given by the model: Positive (green), Negative (red) and Neutral (yellow).} 
\vspace{-2em}
\label{fig:exampleresults}
\end{figure}

\begin{table}[ht!]
\vspace{-2.5em}
\centering
\caption{Performance results achieved at image and event level.}
\label{imageevaluation}
\begin{tabular}{c|c|c|c|cc|c|c|c|cc|}
 & \multicolumn{5}{c|}{Image Classification} & \multicolumn{5}{c|}{Event Classification} \\ \cline{2-11} 
 & Pos & Neg & Neu & \multicolumn{2}{c|}{All} & Pos & Neg & Neu & \multicolumn{2}{c|}{All} \\ \cline{2-11} 
 & \multicolumn{3}{c|}{mean} & mean & std & \multicolumn{3}{c|}{mean} & mean & std \\ \cline{2-11} 
Semantic Features & 59.2 & 42.4 & 44.4 & 48.67 & 22.87 & 71.2 & 42 & 47.3 & 53.50 & 30.77 \\
CNN Features & 70 & 61.3 & 45.7 & 59.00 & 22.80 & 80.8 & 71 & 48.9 & 66.90 & 27.67 \\
Semantic$+$CNN Features  & 72 & 60.8 & 46 & 59.60 & 23.17 & 82.1 & 73.5 & 48.9 & 68.17 & 30.07
\end{tabular}
\vspace{-2.5em}
\end{table}
In Table \ref{imageevaluation}, we report the results achieved by the proposed methods at image and event level. We achieved an average image classification rate of $59.60\% $ with a standard deviation of $23.17$, when we apply the proposed method. The average event classification rate is $68\%$, when the proposed features are employed, which corresponds to $82\%$, $73.5\%$ and $49\%$ for positive, negative and neutral events, respectively. Up to our knowledge, unfortunately, there is no work in the literature on egocentric image sentiment recognition neither event sentiment recognition to compared with. Even the works on image sentiment analysis in conventional images \cite{Campos2015DivingPrediction,Levi2015EmotionPatterns,You2016BuildingBenchmark,Yu2016VisualNetworks} use different datasets and objectives (8 semantic sentiments vs. binary or ternary sentiment values) that make difficult their direct comparison.
Fig. \ref{fig:exampleresults} shows some example results. As can be seen, the algorithm learns to classify events with presence of routine objects into \textit{neutral} events. 
Events wrongly classified as \textit{neutral} are shown in Fig. \ref{fig:exampleresults}(left) and Fig. \ref{fig:exampleresults}(middle). As an example, the last row of Fig. \ref{fig:exampleresults}(left) is classified as \textit{neutral}, probably due to the presence of the \textit{pc} in the image, while it was manually labelled as \textit{positive}, because it shows social interactions. As for Fig. \ref{fig:exampleresults}(left) and Fig. \ref{fig:exampleresults}(right), events were mislabelled as \textit{negative} probably due to the "homogeneity" and "greyness" of the images within the events, e.g. events were considered as \textit{negative} when most of the information in the image corresponded to the asphalt of the road.


\vspace{-1em}
\subsection{Discussion} 
\label{sec:discussion}
\vspace{-0.5em}
Sentiments recognition from an image or a collection of images is a difficult process due to its ambiguity. A challenge in the model construction for sentiment recognition consists in taking into account the bias due to the subjective interpretation of images by different users. Furthermore, the boundaries between neutral/positive and neutral/negative sentiments are not clearly defined. A \textit{neutral} feeling is difficult to interpret. From the results, we observe that \textit{neutral} events are the most challenging ones to classify. Another challenging aspect concerns the grouping of image sentiments into event sentiment, since events can have non-uniform sentiments. 

A further step towards better understanding of the image and sentiment analysis is needed, due to the subjectivity of what an image can recall to different persons. To this aim, having annotations by different persons is critical to evaluate the inter- and intra-observer variability.

From the results, the intuition that we get is that non-routine events and specially when moments are social,  have a higher probability of being positive. In contrast, routine events will most probably be considered as neutral. Negative events as accidents have low prevalence to be learned. Yet, hostile and empty environments could lead to negative sentiments too. Future works will address the study of emotional events and their relation to daily routine.

\vspace{-1em} 

\section{Conclusions} 
\label{sec:conclusions}
\vspace{-0.7em}
In this work, we propose, for the first time, a system and a dataset for egocentric sentiment image and event recognition based on the extraction of CNN and semantic features with sentiment value associated. 
We introduced a new labelled dataset of egocentric images composed of 233 events, grouping 12088 images, from 20 days of 3 users grouped.
We presented preliminary results, obtaining an average events and image sentiment accuracy of 68.17\% and 58.60\%, with std of 30.07\% and 23.17\%, respectively. 
\vspace{-0.5em}
\section*{Acknowledgements}
\vspace{-0.5em}
This work was partially founded by TIN2015-66951-C2, SGR 1219, CERCA, \textit{ICREA Academia'14} and Grant 20141510 (Marat\'{o}TV3). The funders had no role in the study design, data collection, analysis, and preparation of the manuscript. 
\vspace{-0.5em}
\bibliographystyle{abbrv} 
\vspace{-1.7em}
\bibliography{Mendeley.bib}
\end{document}